\documentclass[journal]{IEEEtran}

\usepackage{graphicx} 
\usepackage[colorlinks=true,allcolors=blue]{hyperref}
\usepackage{multirow}
\usepackage{adjustbox}
\usepackage{tabularx}
\usepackage{booktabs}

\begin{document}

\title{A High-Level Survey of Optical Remote Sensing}

\author{

  \IEEEauthorblockN{
    Panagiotis~Koletsis\IEEEauthorrefmark{1},
    Vasilis~Efthymiou\IEEEauthorrefmark{1},
    Maria~Vakalopoulou\IEEEauthorrefmark{2}\IEEEauthorrefmark{5},
    Nikos~Komodakis\IEEEauthorrefmark{3}\IEEEauthorrefmark{5},
    Anastasios~Doulamis\IEEEauthorrefmark{4},
    Georgios~Th.~Papadopoulos\IEEEauthorrefmark{1}\IEEEauthorrefmark{5}\\
    \thanks{This work has received funding from the European Union’s Horizon Europe research and innovation programme under Grant Agreement No. 101168042 project TRIFFID (auTonomous Robotic aId For increasing First responders Efficiency). The views and opinions expressed in this paper are those of the authors only and do not necessarily reflect those of the European Union or the European Commission.}
    }

    \IEEEauthorblockA{
        \IEEEauthorrefmark{1}
        Department of Informatics and Telematics, Harokopio University of Athens, Athens, Greece\\
        \texttt{\{pkoletsis,vefthym,g.th.papadopoulos\}@hua.gr}
     }

    \IEEEauthorblockA{
        \IEEEauthorrefmark{2}
        CentraleSupelec, University Paris Saclay, Paris, France\\
        \texttt{maria.vakalopoulou@centralesupelec.fr}
     }
     
    \IEEEauthorblockA{
        \IEEEauthorrefmark{3}
        Computer Science Department, University of Crete, Heraklion, Greece\\
        \texttt{komod@csd.uoc.gr}
    }

    \IEEEauthorblockA{
        \IEEEauthorrefmark{4}
        Department of Topography, National Technical University of Athens, Athens, Greece\\
        \texttt{adoulam@cs.ntua.gr}
    }

    \IEEEauthorblockA{
        \IEEEauthorrefmark{5}
        Archimedes, Athena Research Center, Athens, Greece\\
    }    
  
}

\maketitle
\begin{abstract}
 In recent years, significant advances in computer vision have also propelled progress in remote sensing. Concurrently, the use of drones has expanded, with many organizations incorporating them into their operations. Most drones are equipped by default with RGB cameras, which are both robust and among the easiest sensors to use and interpret. The body of literature on optical remote sensing is vast, encompassing diverse tasks, capabilities, and methodologies. Each task or methodology could warrant a dedicated survey. This work provides a comprehensive overview of the capabilities of the field, while also presenting key information, such as datasets and insights. It aims to serve as a guide for researchers entering the field, offering high-level insights and helping them focus on areas most relevant to their interests. To the best of our knowledge, no existing survey addresses this holistic perspective.
\end{abstract}


\section{Introduction}\label{sec:introduction}
Earth observation (EO) is the collection and analysis of data about Earth, acquired by sensors and high-resolution cameras of different bandwidths, that are typically placed on satellites, drones, and aircrafts. 
EO is of paramount importance to society, as a plethora of frameworks that contribute to the greater good are based on this concept, particularly in environmental and climate monitoring, including tracking temperature changes, greenhouse gas concentrations, deforestation, desertification, and pollution trends to inform science and policy~\cite{lee2023ipcc}. EO also enables air quality monitoring~\cite{jutz2020copernicus,ramapriyan2009evolution} and supports various agriculture tasks~\cite{kogan2019remote}.


Optical Remote Sensing (ORS) is a core component of EO, predating the satellite era. While satellite imagery often relies on multi- or hyperspectral data, RGB sensors remain the most accessible and widely used modality due to their cost-effectiveness and intuitive natural-color imagery. Their prevalence has been further strengthened by the widespread adoption of affordable and reliable Unmanned Aerial Vehicles (UAVs), most of which are equipped with high-resolution RGB sensors that provide an effective balance between performance and cost~\cite{drones7060398}.

The utility of RGB-based ORS has been amplified by advances in sensing and data infrastructure. Improvements in optics, sensor design, and miniaturization have produced lightweight, high-performance cameras suitable for diverse remote sensing applications~\cite{kim2024metasurface}. In parallel, progress in image and video compression~\cite{rahman2019lossless}, visual processing \cite{rodis2024multimodal, cani2026illicit} and high-speed communication networks~\cite{DBLP:journals/network/SaadBC20} enables efficient storage and near–real-time transmission of large volumes of high-resolution visual data.

Numerous survey papers have been published in recent years in EO-related fields; however, none provide a comprehensive overview of the capabilities and technologies that span the entire domain. Most existing surveys focus on specific tasks rather than sensing modalities~\cite{lv2021land,cheng2024methods}. Some surveys cover the major ORS task~\cite{li2024review,li2022water} or even specialized sub tasks~\cite{li2022cloud,zhang2024development}.
Other surveys adopt a domain-driven perspective, focusing on domain applications~\cite{victor2024systematic}. Several works examine learning strategies in remote sensing~\cite{hosseiny2023beyond,wang2022empirical}. More recent surveys emphasize broader methodological trends, particularly foundation models~\cite{DBLP:journals/corr/abs-2503-22081,11119145}.
Unlike existing surveys that focus on individual tasks, learning paradigms, or application domains, this work adopts a modality-centric perspective, providing a unified overview of RGB-based optical remote sensing across tasks, datasets, and emerging trends. By jointly reviewing tasks, benchmarks, and recent foundation models, the survey offers a practical entry point for researchers working with the most widely available EO imagery in RGB.
The RGB imagery requires no additional domain knowledge to interpret, unlike other spectral bands, which often require geographical or physical expertise. 

\textit{Bibliometrics:}
The search for papers was conducted using two main databases: Elsevier Scopus\footnote{https://www.scopus.com/pages/home\#basic}
and IEEE Xplore\footnote{https://ieeexplore.ieee.org/Xplore/home.jsp}. 
The search focused mainly on the top 20 remote sensing venues provided by Google Scholar and was restricted to the last 4 years [2022,2025]. Popular AI and Computer Vision venues were added as well. The final selection of  articles was based on criteria such as citations, authors, and task diversification.

\textbf{Outline.} The remainder of the paper is organized as follows. Section~\ref{sec:mainORStasks} presents a categorization of the identified ORS tasks. Section~\ref{sec:datasets} analyzes the available datasets for each task. Section~\ref{sec:latestTrends} discusses the latest trends in the field. Section~\ref{sec:insigtsANDtopics} presents the insights and open topics that emerged from this work. Finally, Section~\ref{sec:conclusion} concludes the paper.

\section{Main ORS tasks}\label{sec:mainORStasks}
\begin{figure*}
  \centering
  \includegraphics[width=1\linewidth]{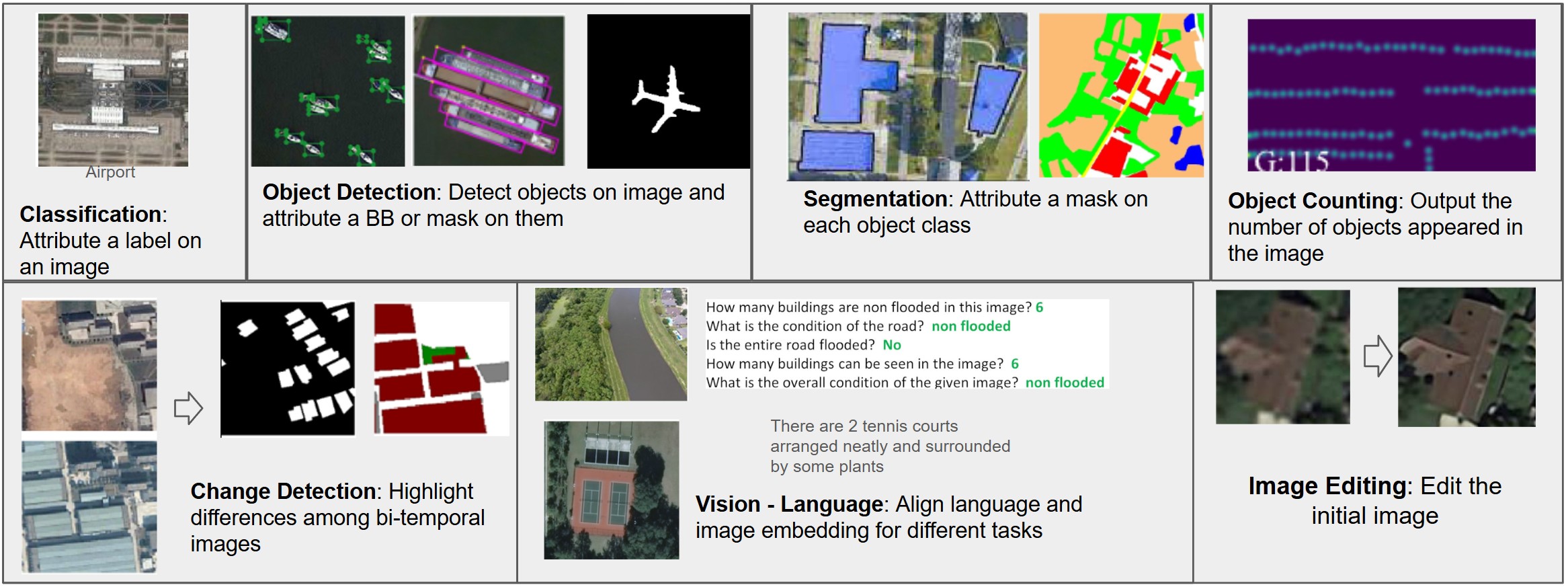}
  \caption{Main tasks in optical remote sensing.}
  \label{fig:Overviewtasks}
\end{figure*}

In this section, we provide an overview of the main tasks in ORS, as also illustrated in Figure~\ref{fig:Overviewtasks}, and their corresponding sub-tasks. 

\subsection{Classification}
Classification is one of the most fundamental tasks in ORS, forming the basis of high-level scene understanding and decision-making. It provides global semantic interpretation of RS imagery, enabling large-scale applications such as land-use/land-cover mapping, urban growth analysis, and disaster assessment.

\textbf{Image/Scene Classification} aims to assign a label to an image. In recent years, the successful adoption of transformer-based~\cite{DBLP:conf/nips/VaswaniSPUJGKP17,DBLP:conf/iclr/DosovitskiyB0WZ21} models in vision has led to extensive exploration of transformer-based approaches for this task. SCViT~\cite{DBLP:journals/tgrs/LvWZDZ22} employs a transformer architecture that captures both local and global image structure, with local features encoding object-level patterns and global features encoding image-level context. Other work~\cite{DBLP:journals/lgrs/DengXH22} proposes a hybrid design that combines CNN~\cite{DBLP:conf/nips/KrizhevskySH12} modules for local feature extraction with transformer modules for global representation, with predictions derived from both. Despite the rise of transformers, purely CNN-based models remain relevant due to their maturity; for example, MGSNet~\cite{DBLP:journals/tgrs/WangLZTC23} relies exclusively on CNN modules throughout the network.

\textbf{Cross-Scene Classification} extends classical image/scene classification by training and testing on different datasets, emphasizing generalization. HFPAN~\cite{DBLP:journals/tgrs/MiaoHGJ24} leverages fine-grained and contextual information via global and local correlations, using hierarchical feature embeddings and alignment to achieve source domain adaptation across datasets.

\textbf{Fine-Grained Classification} is a variant of image/scene classification focusing on narrow domains where visual differences are subtle and inter-class similarity is high. For example, the broad category of airplanes may include subcategories such as Boeing 747 and Airbus A330. RIFCRN~\cite{DBLP:journals/tgrs/LiC0W24} proposes a CNN-based few-shot approach for fine-grained vehicle classification.

\subsection{Object Detection}
Object Detection (OD) is a critical task in ORS, aiming to localize and categorize objects of interest within the image, thus enabling more detailed and instance level scene understanding compared to image classification.

\textbf{Horizontal Object Detection} (HOD) aims to localize objects of interest by generating one or more horizontal bounding boxes, each associated with a class label describing the object within the region. The YOLO~\cite{DBLP:conf/cvpr/RedmonDGF16} architecture has gained significant attention due to its ease of training and competitive performance. For instance, LAR-YOLOv8~\cite{DBLP:journals/staeors/YiLZL24} enhances YOLOv8 by integrating a transformer block into the CNN-based feature extractor to improve small-object detection. Other works focus on efficiency in HOD, such as CFANet~\cite{DBLP:journals/tgrs/ZhangWGZL23} and CEASC~\cite{DBLP:conf/cvpr/DuHC023}, which are pure CNN-based models optimized for detecting small objects like vehicles and humans.

\textbf{Oriented Object Detection} extends horizontal object detection by employing rotated bounding boxes that incorporate orientation angles. These oriented boxes provide tighter localization, making OOD well suited for densely packed objects and complex occlusions. Similar to HOD, most OOD approaches are purely CNN-based, integrating feature alignment and anchor refinement modules~\cite{DBLP:journals/tgrs/MingMZD22,DBLP:journals/tgrs/ChengYLLXWYH22,DBLP:journals/tgrs/HanDLX22}. 

\textbf{Fine-Grained OD} is a variant of standard OD, just like in classification, the instances of interest are from a specific class. This category is possible in both HOD~\cite{DBLP:journals/tgrs/HanYPP22} and OOD setup~\cite{Li18082022}.

\textbf{Salient Object Detection} aims to identify an object in the image but instead of producing a bounding box on the object, it produces a binary segmentation mask (see the airplane mask in Figure~\ref{fig:Overviewtasks}). An interesting observation across these works is that most methods are exclusively CNN-based. They typically rely on multiscale feature integration to address object scale variability and edge refinement modules to avoid blurry or coarse boundaries, thereby improving salient object detection accuracy in remote sensing images~\cite{9631225,9474908}. Although some hybrid approaches are also observed~\cite{9791375}.

\textbf{Video Object Tracking} (VOT) extends horizontal object detection (HOD) to video data, where temporally connected frames make time a critical factor. The large number of frames poses scalability challenges due to increased computational complexity. CFDRM~\cite{10535900} investigates a purely CNN-based approach for weakly supervised VOT.

\subsection{Segmentation}
Segmentation in ORS aims to provide detailed, pixel-level understanding of imagery, offering finer granularity compared to classification or object detection.

\textbf{Semantic Segmentation (SS)} focuses on scene-level understanding by assigning each pixel to a semantic class, enabling precise localization but without distinguishing individual object instances. A background class is typically included to suppress noise. Various architectures have been tested for the domain, from pure CNN~\cite{WIELAND2023113452,JIANG202410} that combine efficiency with effectiveness to hybrid approaches~\cite{fu2024complementarity,dai2023radanet}. Other works explore the effectiveness of the SAM foundation model for semantic segmentation~\cite{osco2023segment,10990024}. Finally, some approaches leverage generative models—such as Generative Adversarial Network (GANs)~\cite{Wang31122022,zhang2021stagewise} and diffusion models~\cite{ayala2023diffusion}—for domain adaptation or to enhance mask quality.

\textbf{Instance Segmentation (IS)} focuses on identifying individual object instances and generating a segmentation mask for each, including occluded objects of the same class. Similar to semantic segmentation, most approaches are CNN-based. For example~\cite{AMOBOATENG2022569} introduce a roof instance segmentation dataset from Ghana and propose a purely CNN-based method using ResNet as the feature extractor and Mask R-CNN~\cite{DBLP:conf/iccv/HeGDG17} as the segmentation head. Hybrid approaches have also emerged: Chen et al.~\cite{CHEN2023129} present an improved hybrid Mask R-CNN~\cite{DBLP:conf/iccv/HeGDG17} incorporating a transformer module and a super-resolution component to enhance image quality.

\subsection{Change Detection}
Change Detection (CD) in ORS identifies temporal variations on the Earth’s surface from bi-temporal images, providing fine-grained spatiotemporal information for applications such as disaster assessment, environmental monitoring, infrastructure inspection, and agricultural or forestry management. CD is challenging due to image misalignment, seasonal variations, and the difficulty of detecting small changes, which also makes dataset annotation slow and costly even for experts.

\textbf{Binary Change Detection (BCD)} focuses on identifying binary changes by producing a segmentation mask. A2Net~\cite{DBLP:journals/tgrs/LiTLZDWZ23} is a lightweight CNN using MobileNetV2 as a feature extractor. BiFA~\cite{DBLP:journals/tgrs/ZhangCZCLZS24} is a pure Transformer-based lightweight method, while the approaches proposed in~\cite{DBLP:journals/tgrs/LiCDWWLP24,DBLP:journals/tgrs/ChenQS22} are hybrid models. Recent works, such as MambaCD~\cite{DBLP:journals/tgrs/ChenSHXY24}, adopt the novel Mamba architecture.

\textbf{Semantic Change Detection (SCD)} extends change detection by assigning semantic labels to different types of changes in bi-temporal images, producing a multiclass segmentation mask. J. Ge et al.~\cite{Ge31122024} proposed a CNN-based real-time system to classify buildings affected by earthquakes. HGINet~\cite{LONG2024318} uses a transformer-based architecture as the backbone and a Graph Convolutional Network (GCN)-based prediction head, providing strong generalization ability and robustness.

\subsection{Vision-Language}
Vision-Language tasks in ORS aim to bridge visual and textual information, enabling models to interpret imagery in a semantically rich, human-understandable way. These models allow users to interact via simple textual prompts without technical expertise. The use of a non-domain-specific vocabulary has also enabled foundation models, which develop general understanding and broad applicability across tasks.


\textbf{Image Captioning} aims to generate textual descriptions of images for global understanding and interpretation. For example, BITA \cite{DBLP:journals/tgrs/YangLZ24} adopts a two-stage framework: (1) vision–language pretraining to align image–text features using a lightweight interactive Fourier transform, and (2) caption generation via an LLM guided by the aligned features from a frozen encoder.

\textbf{Vision Question Answering (VQA)} enables models to answer text-based questions about an image, facilitating interactive user–model engagement. CDVQA~\cite{DBLP:journals/tgrs/YuanMXZ22} introduces VQA specifically for CD images.

\textbf{Visual Grounding} is similar to object detection but incorporates a textual description of the target object to produce a bounding box. RSVG~\cite{DBLP:journals/tgrs/ZhanXY23} introduces this task for remote sensing images, providing a dataset based on DIOR (Section~\ref{sec:datasets}) along with a benchmark.

\subsection{Image/Video Editing}
Image{/}Video Editing in ORS focuses on enhancing, restoring, or transforming visual content to improve usability, interoperability, and analysis of remote sensing data.


\textbf{Image Super-Resolution (ISR)} aims to increase the spatial resolution of an image, enhancing its visual clarity and perceptual quality. Various architectures have already been tested, ranging from hybrid models~\cite{DBLP:conf/igarss/SuiMZP23} to pure transformers~\cite{DBLP:journals/tgrs/LeiSM22}. Recent works in the field also explore the effectiveness of diffusion probabilistic models (DPMs)~\cite{DBLP:journals/tgrs/ZhangLLGSJ24}. EDiffSR~\cite{DBLP:journals/tgrs/XiaoYJHJZ24} leverages the strengths of DPMs for high-quality image generation with an efficient approach, employing two main modules: one for efficient noise prediction and another for extracting enriched conditions.


\textbf{Video Super Resolution (VSR)} is closely related to ISR, but operates on a sequence of images rather than a single image, i.e., a video~\cite{DBLP:journals/tgrs/XiaoSYLSZ22,DBLP:journals/aeog/XiaoYHZSSWZ22}. Consequently, handling high frame rates becomes challenging, and computational efficiency emerges as a key constraint.

\subsection{Object Counting}
Object counting (OC) in ORS aims to estimate the number of object instances within an image, providing critical quantitative information for applications such as traffic monitoring, crowd analysis, vehicle and ship counting, and agricultural or forestry assessment.

\textbf{Single Object Counting (SOC)} aims to identify and count all the same objects of a specific class appearing in the provided image. HKINet~\cite{DBLP:journals/tgrs/WangZZWHWZ24} extracts semantic features while preserving high-resolution details. The hierarchical features from multiresolution convolutions are then fused to generate the final prediction density map.


\textbf{Multiclass Object Counting (MOC)} acts similarly to the SOC with the difference that the output is more than one number, i.e one for each class of objects. DOPNet~\cite{DBLP:journals/tgrs/CuiDYC24} uses a localization module to convert SOC into MOC, predicts multiple density maps, and assigns category attributes to each object. It leverages point-level annotations to learn pseudo-box predictions for every object.

\subsection{Other Tasks}
Other tasks include those beyond the typical ORS tasks described above. Often specific to particular scenarios or originally developed for non-RGB images.

\textbf{Geo-localization} aims to match a satellite picture to another image; this image could have been acquired from drone~\cite{DBLP:journals/tgrs/WangSQJLS0Q24} or even from Point Of View (POV) field~\cite{DBLP:conf/cvpr/ZhuS022}.

\textbf{Accident Prediction} task takes as input an ORS image and outputs some point-dots with high risk of a potential car accident~\cite{DBLP:journals/staeors/LiangZXJ24}. This task is feasible because the authors proposed a dataset specifically for it. Using a self-supervised approach followed by supervised fine-tuning, the model learned the potential risks associated with each area of the road.

\textbf{Canopy Height (CH)} task is typically a lidar task transferred to only RGB bands. This task produces a mask which each pixel has a color representing the corresponding height level of the surface~\cite{TOLAN2024113888}.

\textbf{Image compression} compresses an image without missing critical information and without significantly reducing the resolution~\cite{DBLP:journals/tgrs/XiangL24}. This is particularly useful because ORS images are typically very large and require substantial storage.

\begin{table*}[ht!]
  \caption{Datasets for ORS Tasks. SRes:~Spatial Resolution of the image in meters. \#Expr:~number of matching expressions to the images. Y:~Year of publication. HMO: Human-Made Objects. `-':~Defines that this information was not provided in the paper.}
  \label{tab:R2}
  \centering

  \begin{tabular}{cccccccc}
    \toprule
    \multicolumn{8}{c}{\textbf{Classification}} \\
    Name & size & domain & \#classes & SRes & Source & Image Size & Y\\
    \midrule
    UCM~\cite{DBLP:conf/gis/YangN10} & 2100 & Construction \& Nature & 21 & 0.3 & Satellite  & 256x256 & 2010\\
    AID~\cite{DBLP:journals/tgrs/XiaHHSBZZL17} & 10000 & Construction \& Nature & 30 & 0.5-8 & Satellite  & 600x600 & 2017\\
    NWPU-RESISC45~\cite{DBLP:journals/pieee/ChengHL17}  & 31500 & Construction \& Nature & 45 & 0.2-30 & Satellite & 256x256 & 2017\\
    \bottomrule
  \end{tabular}
  
  \vspace{1em} 

  \begin{tabular}{cccccccccc}
    \toprule
    \multicolumn{10}{c}{\textbf{Object Detection}} \\
    Name & subtask & Size & Domain & \#classes & \#instances & SRes & Source & Image Size & Y\\
    \midrule
     NWPU VHR-10~\cite{CHENG2014119} & Horizontal & 715 & HMO & 10 & 3775 & 0.5-2 & Satellite & - & 2014\\
     DOTA-v.1~\cite{DBLP:conf/cvpr/XiaBDZBLDPZ18} & Oriented & 2806 & HMO & 15 & 188282 & 0.1 - 4.5 & Satellite & $256^2$ - 4000x4000 & 2018\\
     LEVIR~\cite{DBLP:journals/tip/ZouS18} & Horizontal & 21952 & HMO & 3 & 11028 & 0.2-1 & Satellite & 600x800 & 2018\\
     ORSSD~\cite{DBLP:journals/tgrs/LiCHZQK19} & Salient & 800 & HMO \& Nature & 8 & - & 0.5-2 & Satellite & $256^2$ - 1264x987 & 2019 \\
     DIOR~\cite{DBLP:journals/corr/abs-1909-00133} & Horizontal & 23463 & HMO & 20 & 192472 & 0.5-30 & Satellite & 800x800 & 2020\\
     EORSSD~\cite{DBLP:journals/tip/ZhangCLCFCZK21} & Salient & 2000 & HMO \& Nature & 8 & - & 0.5-3 & Satellite & $256^2$ - 1264x988 & 2020\\
     DOTA-v.2~\cite{DBLP:journals/pami/DingXXBYYBLDPZ22} & Oriented & 11268 & HMO & 18 & 1793658 & 0.1 - 4.5 & Satellite & $800^2$ - 29200x27620 & 2021\\
     VisDrone~\cite{DBLP:conf/iccvw/CaoHWWYZZZGHHHL21} & Horizontal & 10209 & Vehicles \& Humans & 10 & - & - & UAV & - & 2021\\
     FAIR1M~\cite{DBLP:journals/corr/abs-2103-05569} & Oriented & 40000 & HMO & 37 & 1000000 & 0.3-8 & Satellite & - & 2022\\
     DIOR-R~\cite{DBLP:journals/tgrs/ChengWLXLYH22} & Oriented & 23463 & HMO & 20 & 192472 & 0.5-30 & Satellite & 800x800 & 2022\\
     
    \bottomrule
  \end{tabular}

  \vspace{1em}
  
  \begin{tabular}{cccccccccc}
  \toprule
    \multicolumn{10}{c}{\textbf{Segmentation}} \\
    Name & subtask & Size & Domain & \#classes & \#instances & SRes& Source & Image Size & Y\\
    \midrule
    Inria~\cite{DBLP:conf/igarss/MaggioriTCA17} & Semantic & 360 & Construction & 2 & - & 0.3 & Aerial  & 1500x1500 & 2017\\
    iSAID~\cite{DBLP:conf/cvpr/ZamirAGKSK00XB19} & Instance & 2806 & HMO & 15 & 655451 & 0.1-4.5 & Satellite  & $800^2$ - $4000^2$ & 2019\\
    WHU-Building~\cite{DBLP:journals/tgrs/JiWL19} & Semantic & 8189 & Construction & 2 & 220000 & 0.075- 2.7 & Satellite  & 512x512 & 2019\\
    LoveDA~\cite{DBLP:conf/nips/WangZMLZ21} & Semantic & 5987 & Construction \& Nature & 7 & 166768 & 0.3 & Satellite  & 1024x1024 & 2019\\
    \bottomrule
  \end{tabular}

  \vspace{1em}

  \begin{tabular}{cccccccccc}
  \toprule
    \multicolumn{10}{c}{\textbf{Object Counting}} \\
    Name & subtask & Size & Domain & \#classes & \#instances & SRes & Source & Image Size & Y\\
    \midrule
    RSOC~\cite{DBLP:journals/tgrs/GaoLW21} & Single & 3057 & HMO & 4 & 286539 & 0.1-4.5 & Satellite  & $256^2$ - 4000x4000 & 2021\\
    DroneCrowd~\cite{DBLP:conf/cvpr/WenDZHWBL21} & Single & 33960 & Humans & 1 & 4800000 & - & UAV  & 1920x1080 & 2021\\
    \bottomrule
  \end{tabular}

  \vspace{1em}
  
  \begin{tabular}{cccccccccc}
  \toprule
    \multicolumn{10}{c}{\textbf{Change Detection}} \\
    Name & subtask & Size(pairs) & Domain & \#classes & \#instances & SRes& Source & Image Size & Y\\
    \midrule
    CDD~\cite{isprs-archives-XLII-2-565-2018} & Binary & 16000 & HMO \& Nature & 2 & - & 0.003 - 1 & Satellite  & 256x256 & 2018\\
    WHU-CD~\cite{DBLP:journals/tgrs/JiWL19} & Binary & 7620 & Construction & 2 & 3381 & 0.3 & Satellite  & 512x512 & 2019\\
    LEVIR-CD~\cite{DBLP:journals/remotesensing/ChenS20} & Binary & 637 & Construction & 2 & 31000 & 0.5 & Satellite  & 1024x1024  & 2020\\
    Second~\cite{DBLP:journals/tgrs/YangXLDYPZ22} & Semantic & 4662 & Construction \& Nature & 7 & - & - & Satellite  & 512x512 & 2021\\
    SYSU-CD~\cite{DBLP:journals/tgrs/ShiLLLWZ22} & Binary & 20000 & Construction & 2 & - & 0.5 & Satellite  & 256x256 & 2021\\
    S2Looking~\cite{DBLP:journals/remotesensing/ShenLCWXYCLJ21}  & Binary & 5000 & Construction & 2 & 65920 & 0.5 - 0.8 & Satellite  & 1024x1024 & 2021\\
    Landsat-SCD~\cite{DBLP:journals/digearth/YuanZZWLZ22}& Semantic & 8468 & Construction \& Nature & 10 & - & 30 & Satellite  & 416x416  & 2022\\
    
    \bottomrule
  \end{tabular}
  \vspace{1em}

  \begin{tabular}{cccccccccc}
  \toprule
    \multicolumn{10}{c}{\textbf{Vision - Language}} \\
    Name & subtask & Size & Domain & \#classes & \#Expr & SRes & Source & Image Size & Y\\
    \midrule
    DIOR-RSVG~\cite{DBLP:journals/tgrs/ZhanXY23} & Visual Grounding & 17402 & HMO & 20 & 38320 & 0.5-30 & Satellite  & 800x800 & 2023 \\
    RS5M~\cite{DBLP:journals/tgrs/ZhangZGY24} & Image Captioning & 5000000 & - & - & - & - & Satellite  & - & 2024\\
    CDVQA~\cite{DBLP:journals/tgrs/YuanMXZ22} & CDVQA & 4662 & Construction \& Nature & - & - & 0.5-3 & Satellite  & 512x512 & 2024 \\
    \bottomrule
  \end{tabular}

\end{table*}


\section{Datasets}\label{sec:datasets}

This section describes the most popular datasets for each task in the ORS domain. All presented datasets are publicly available and can be found in Table~\ref{tab:R2}.

\textit{Classification} datasets are among the oldest in the field, as classification is the simplest entry point. In all 3 datasets, methods have achieved highly competitive results and they are composed of elements from construction and nature domain. \textit{Object detection} is the second oldest task that adds localization to objects. Datasets in this domain are typically very large, containing many instances, and are therefore suitable for large-scale training for both horizontal object detection (DIOR~\cite{DBLP:journals/corr/abs-1909-00133} and DOTA v1~\cite{DBLP:conf/cvpr/XiaBDZBLDPZ18}) and oriented object detection (DOTA v2~\cite{DBLP:journals/pami/DingXXBYYBLDPZ22} and FAIR1M~\cite{DBLP:journals/corr/abs-2103-05569}). VisDrone~\cite{DBLP:conf/iccvw/CaoHWWYZZZGHHHL21} is a UAV dataset for urban containing classes as vehicles and humans. In addition, salient object detection datasets are significantly smaller on images, classes, and instances compared to the other two categories. \textit{Segmentation} datasets are focusing on construction elements, such datasets are Inria~\cite{DBLP:conf/igarss/MaggioriTCA17}, WHU-Buildings~\cite{DBLP:journals/tgrs/JiWL19} and LoveDA~\cite{DBLP:conf/nips/WangZMLZ21}, with the third having nature elements as well. iSAID~\cite{DBLP:conf/cvpr/ZamirAGKSK00XB19} is a large scale dataset for instance segmentation for Human Made Objects (HMO) domain. \textit{Object counting} datasets are limited in number and are suitable for single object detection. \textit{Change Detection} datasets are among the most challenging to create, and as a result, they are typically smaller than those for other tasks. Most binary change detection datasets focus on the construction domain (WHU-CD~\cite{DBLP:journals/tgrs/JiWL19}, LEVIR-CD~\cite{DBLP:journals/remotesensing/ChenS20}, S2Looking~\cite{DBLP:journals/remotesensing/ShenLCWXYCLJ21}). For semantic change detection, datasets include both construction and natural changes, as these are the most important changes to track. \textit{Vision-Language} datasets are the most recently introduced. They often reuse images from large-scale datasets like DOTA~\cite{DBLP:journals/pami/DingXXBYYBLDPZ22} and FAIR1M~\cite{DBLP:journals/corr/abs-2103-05569}, adding textual descriptions or prompts to define new benchmarks, reducing the need for new image collection. These datasets are valuable for evaluating foundation models that integrate multimodality and multitasking. Tools like large language models (LLMs) and vision–language models (VLMs) can simplify dataset creation through automation. Notable examples are RS5M~\cite{DBLP:journals/tgrs/ZhangZGY24} for captioning, DIOR-RSVG~\cite{DBLP:journals/tgrs/ZhanXY23} for visual grounding and CDVQA~\cite{DBLP:journals/tgrs/YuanMXZ22} for CD question answering.

\section{Latest Trends}\label{sec:latestTrends}
The latest trend observed in the field is the emergence of foundation models (FM). This trend highlights the need for a single model capable of performing multiple tasks. Such capability is achieved through the use of large-scale models, pre-trained on vast image datasets, most commonly in a self-supervised \cite{konstantakos2025self} manner. Subsequently, supervised fine-tuning is performed on downstream tasks.

Several major works in computer vision have emerged lately, such as Grounding DINO~\cite{DBLP:conf/eccv/LiuZRLZYJLYSZZ24}, SAM~\cite{DBLP:conf/iccv/KirillovMRMRGXW23}
etc. These advancements and trends have been incorporated into RS over the past few years. This occurs either through adaptation~\cite{DBLP:journals/tgrs/DingZPTYB24} or by developing a new model trained on RS imagery that is specifically tailored to the domain.
SMLFR~\cite{DBLP:journals/tgrs/DongGL24} is a CNN based RS-FM pre-trained on 9 million RS images using a Self Supervised Learning approach. The model is able to perform OOD, SS, and CD by employing three separate heads, each trained in a supervised manner on popular datasets.
RingMo~\cite{DBLP:journals/tgrs/SunWLZLHLRYCHYWLF23} is a transformer-based FM trained similarly to SMLFR, using Self Supervised Learning pretraining and supervised finetuning, capable of performing classification, OD, SS, CD. 
RemoteCLIP~\cite{DBLP:journals/tgrs/LiuCGZZYFZ24} is another FM offering 3 versions, 1 with ResNet backbone and 2 with ViT backbone; the ViT based one being superior. This model is capable of classification, image captioning, and object counting. The CLIP model~\cite{DBLP:conf/icml/RadfordKHRGASAM21} is responsible for the alignment of image features with textual phrases.
GeoRSCLIP~\cite{DBLP:journals/tgrs/ZhangZGY24} is once again a transformer based model utilizing the CLIP method for classification, image to text, text to image and semantic localization.

\section{Insights and Open Topics}\label{sec:insigtsANDtopics}
In this section, the insights and open topics produced by this work are presented, as well as a small conclusion is also presented. The insights and trends were primarily derived from the state-of-the-art (SOTA) analysis summarized in Table~\ref{tab:metrics}, where SOTA results are reported for the most commonly used dataset in each task.

\textit{Insights:}
Overall, the surveyed literature highlights that no single architecture is universally optimal across all ORS tasks. CNNs consistently demonstrate strong performance in scenarios dominated by local patterns, such as homogeneous image classification, small-object detection, object counting, and local change detection, while also offering superior computational efficiency. In contrast, Transformer-based models excel in handling heterogeneous scenes and tasks that require global context modeling, including complex object detection, segmentation, and vision–language alignment, albeit at a higher computational cost. As a result, hybrid architectures that combine CNNs and Transformers emerge as a balanced and increasingly dominant solution, achieving strong performance across diverse datasets and tasks by leveraging complementary strengths.

Task-specific insights further reveal clear architectural preferences shaped by data characteristics and annotation constraints. Segmentation and change detection benefit from hybrid designs due to the need for both fine-grained detail and global structure, while image and video editing favor lightweight CNN-based approaches for efficiency. Vision–language tasks naturally align with Transformer-based designs, and adversarial settings expose trade-offs between GAN efficiency and diffusion model stability. Collectively, these findings underscore the importance of selecting architectures based on task requirements, data availability, and efficiency constraints, rather than seeking a single universally superior model architecture.

\begin{table}[t]
\centering
\caption{Per task state-of-the-art performance the most popular public datasets.}
\label{tab:metrics}
\resizebox{\columnwidth}{!}{%
\begin{tabular}{c c c c c}
\toprule
Task & Dataset & Metric & Method & Value \\
\midrule

\multirow{6}{*}{Classification}
& \multirow{6}{*}{NWPU-RESIC45~\cite{DBLP:journals/pieee/ChengHL17}}
& \multirow{6}{*}{OA}
& MGDNet~\cite{DBLP:journals/tgrs/MiaoGJ23} & 91.41 \\
& & & GeRSP~\cite{DBLP:journals/tgrs/HuangZGLW24} & 92.74 \\
& & & MGSNet~\cite{DBLP:journals/tgrs/WangLZTC23} & 94.57 \\
& & & SCViT~\cite{DBLP:journals/tgrs/LvWZDZ22} & 94.66 \\
& & & RingMo-Lite~\cite{DBLP:journals/tgrs/WangZZHWNCCZWWS24} & 95.02 \\
& & & CTNET~\cite{DBLP:journals/lgrs/DengXH22} & 95.49 \\
\midrule

\multirow{4}{*}{Horizontal OD}
& \multirow{4}{*}{DIOR~\cite{DBLP:journals/corr/abs-1909-00133}}
& \multirow{4}{*}{mAP 50}
& FSoD-Net~\cite{DBLP:journals/tgrs/WangZCLZLDS22} & 71.80 \\
& & & RSADet~\cite{DBLP:journals/tgrs/YuJ22} & 72.20 \\
& & & GIEM+MES+ECDD~\cite{DBLP:journals/tgrs/GaoLWCNL24} & 73.20 \\
& & & CoF-Net~\cite{DBLP:journals/tgrs/ZhangLW23} & 75.80 \\
\midrule

\multirow{8}{*}{Oriented OD}
& \multirow{8}{*}{DOTA V1~\cite{DBLP:conf/cvpr/XiaBDZBLDPZ18}}
& \multirow{8}{*}{mAP 50}
& CFC-Net~\cite{DBLP:journals/tgrs/MingMZD22} & 73.50 \\
& & & ARS-DETR~\cite{DBLP:journals/tgrs/ZengCYLY24} & 75.47 \\
& & & S2ANet~\cite{DBLP:journals/tgrs/HanDLX22} & 79.42 \\
& & & AOPG~\cite{DBLP:journals/tgrs/ChengWLXLYH22} & 80.66 \\
& & & DODNet~\cite{DBLP:journals/tgrs/ChengYLLXWYH22} & 80.66 \\
& & & -~\cite{DBLP:journals/tgrs/YaoCWLZXH23} & 81.00 \\
& & & FRIOU~\cite{DBLP:journals/tgrs/QianW0Y0H23} & 81.02 \\
& & & RVSA~\cite{DBLP:journals/tgrs/WangZXZDTZ23} & 81.24 \\
\midrule

\multirow{2}{*}{Semantic Segmentation}
& \multirow{2}{*}{LoveDA~\cite{DBLP:conf/nips/WangZMLZ21}}
& \multirow{2}{*}{mIoU}
& GeRSP~\cite{DBLP:journals/tgrs/HuangZGLW24} & 50.56 \\
& & & EMRT~\cite{xiao2023enhancing} & 50.89 \\
\midrule

\multirow{3}{*}{Instance Segmentation}
& \multirow{3}{*}{WHU-Building~\cite{DBLP:journals/tgrs/JiWL19}}
& \multirow{3}{*}{mIoU}
& MRANet~\cite{JIANG202410} & 90.59 \\
& & & RSM-SS~\cite{DBLP:journals/tgrs/ZhaoCZXBO24} & 90.81 \\
& & & UANet~\cite{10416252} & 92.15 \\
\midrule

\multirow{3}{*}{Binary CD}
& \multirow{3}{*}{s2looking~\cite{DBLP:journals/remotesensing/ShenLCWXYCLJ21}}
& \multirow{3}{*}{mIoU}
& SAM-CD~\cite{DBLP:journals/tgrs/DingZPTYB24} & 48.29 \\
& & & BAN-CF~\cite{DBLP:journals/tgrs/LiCM24} & 50.04 \\
& & & TransUNetCD~\cite{DBLP:journals/tgrs/LiZDD22} & 54.41 \\
\midrule

\multirow{4}{*}{Semantic CD}
& \multirow{4}{*}{second~\cite{DBLP:journals/tgrs/YangXLDYPZ22}}
& \multirow{4}{*}{mIoU}
& HGINet~\cite{LONG2024318} & 70.76 \\
& & & ScanNet~\cite{ding2024joint} & 73.42 \\
& & & ChangeMamba~\cite{DBLP:journals/tgrs/ChenSHXY24} & 73.68 \\
& & & FDINet~\cite{DBLP:journals/tgrs/TangFZFSLT24} & 74.35 \\

\bottomrule
\end{tabular}%
}
\end{table}

\textit{Open Research Areas:}
Overall, several open research gaps remain in the ORS domain, including the adaptation of foundation models for multi-modal and multi-task learning. Although FMs are the clear emerging trend and attention has already shifted toward them, most foundation models are not yet competitive with fully supervised training on task-specific datasets. Bridging this gap remains an open challenge, and addressing it would enable foundation models to become a go-to solution across tasks and use cases, helping researchers build robust applications for diverse scenarios.
In addition, the development of efficient diffusion models \cite{alimisis2025advances} for video, and the exploration of video object tracking with oriented bounding boxes. Additional challenges include reviving salient object detection with modern architectures, improving small-object detection, extending Mamba-based models to a wider range of tasks, and systematically evaluating RGB versus multi-/hyper-spectral approaches. Emerging directions such as semantic change detection, learning with limited annotations, domain adaptation, multiclass object counting, dataset design for nature-related objects, standardized editing datasets, as well as efficiency, robustness, and adversarial resilience, collectively highlight the need for scalable, generalizable, and application-ready ORS solutions.

\section{Conclusion}\label{sec:conclusion}
This survey provides a high-level overview of optical remote sensing with a focus on RGB satellite and UAV imagery, emphasizing the capabilities of the domain through a structured presentation of the main tasks and publicly available datasets. In addition, key insights and open research directions were discussed, highlighting current trends and remaining challenges in the field. As RGB imagery continues to dominate satellite and UAV platforms, future progress will increasingly depend on scalable, generalizable, and efficient learning frameworks.

\bibliographystyle{IEEEtran}
\bibliography{refs}

@article{DBLP:journals/tgrs/LvWZDZ22,
  author       = {Pengyuan Lv and others},
  title        = {SCViT: {A} Spatial-Channel Feature Preserving Vision Transformer for Remote Sensing Image Scene Classification},
  journal      = {{IEEE} TGRS},
  year         = {2022},
  doi          = {10.1109/TGRS.2022.3157671},
}

@article{DBLP:journals/tgrs/MiaoGJ23,
  author       = {Wang Miao and others},
  title        = {Multigranularity Decoupling Network With Pseudolabel Selection for Remote Sensing Image Scene Classification},
  journal      = {{IEEE} TGRS},
  year         = {2023},
  doi          = {10.1109/TGRS.2023.3244565},
}

@article{DBLP:journals/tgrs/WangZZHWNCCZWWS24,
  author       = {Yuelei Wang and others},
  title        = {RingMo-Lite: {A} Remote Sensing Lightweight Network With CNN-Transformer Hybrid Framework},
  journal      = {{IEEE} TGRS},
  year         = {2024},
  doi          = {10.1109/TGRS.2024.3360447},
}

@article{DBLP:journals/tgrs/XiaHHSBZZL17,
  author       = {Gui{-}Song Xia and others},
  title        = {{AID:} {A} Benchmark Data Set for Performance Evaluation of Aerial Scene Classification},
  journal      = {{IEEE} TGRS},
  year         = {2017},
  doi          = {10.1109/TGRS.2017.2685945},
}

@inproceedings{DBLP:conf/gis/YangN10,
  author       = {Yi Yang and others},
  title        = {Bag-of-visual-words and spatial extensions for land-use classification},
  booktitle    = {18th {ACM} {SIGSPATIAL} International Symposium on Advances in Geographic Information Systems},
  year         = {2010},
  doi          = {10.1145/1869790.1869829},
}

@article{DBLP:journals/pieee/ChengHL17,
  author       = {Gong Cheng and others},
  title        = {Remote Sensing Image Scene Classification: Benchmark and State of the Art},
  journal      = {Proc. {IEEE}},
  year         = {2017},
  doi          = {10.1109/JPROC.2017.2675998},
}

@article{DBLP:journals/corr/abs-2103-05569,
  author       = {Xian Sun and others},
  title        = {{FAIR1M:} {A} Benchmark Dataset for Fine-grained Object Recognition in High-Resolution Remote Sensing Imagery},
  journal      = {CoRR},
  year         = {2021},
}

@inproceedings{DBLP:conf/cvpr/XiaBDZBLDPZ18,
  author       = {Gui{-}Song Xia and others},
  title        = {{DOTA:} {A} Large-Scale Dataset for Object Detection in Aerial Images},
  booktitle    = {{IEEE} CVPR},
  year         = {2018},
  doi          = {10.1109/CVPR.2018.00418},
}

@article{DBLP:journals/pami/DingXXBYYBLDPZ22,
  author       = {Jian Ding and others},
  title        = {Object Detection in Aerial Images: {A} Large-Scale Benchmark and Challenges},
  journal      = {{IEEE} TPAMI},
  year         = {2022},
  doi          = {10.1109/TPAMI.2021.3117983},
}

@article{DBLP:journals/corr/abs-1909-00133,
  author       = {Ke Li and others},
  title        = {Object Detection in Optical Remote Sensing Images: {A} Survey and {A} New Benchmark},
  journal      = {CoRR},
  year         = {2019},
}

@article{CHENG2014119,
  author       = {Gong Cheng. and others},
  title        = {Multi-class geospatial object detection and geographic image classification based on collection of part detectors},
  journal      = {ISPRS J. Photogramm. Remote Sens.},
  year         = {2014},
  doi          = {10.1016/j.isprsjprs.2014.10.002},
}

@inproceedings{DBLP:conf/iccvw/CaoHWWYZZZGHHHL21,
  author       = {Yaru Cao and others},
  title        = {VisDrone-DET2021: The Vision Meets Drone Object detection Challenge Results},
  booktitle    = {{IEEE/CVF} ICCVW},
  year         = {2021},
  doi          = {10.1109/ICCVW54120.2021.00319},
}

@article{DBLP:journals/tip/ZouS18,
  author       = {Zhengxia Zou and others},
  title        = {Random Access Memories: {A} New Paradigm for Target Detection in High Resolution Aerial Remote Sensing Images},
  journal      = {{IEEE} TIP},
  year         = {2018},
  doi          = {10.1109/TIP.2017.2773199},
}

@article{DBLP:journals/tgrs/LiCHZQK19,
  author       = {Chongyi Li and others},
  title        = {Nested Network With Two-Stream Pyramid for Salient Object Detection in Optical Remote Sensing Images},
  journal      = {{IEEE} TGRS},
  year         = {2019},
  doi          = {10.1109/TGRS.2019.2925070},
}

@article{DBLP:journals/tip/ZhangCLCFCZK21,
  author       = {Qijian Zhang and others},
  title        = {Dense Attention Fluid Network for Salient Object Detection in Optical Remote Sensing Images},
  journal      = {{IEEE} TIP},
  year         = {2021},
  doi          = {10.1109/TIP.2020.3042084},
}

@inproceedings{DBLP:conf/cvpr/ZamirAGKSK00XB19,
  author       = {Syed Waqas Zamir and others},
  title        = {iSAID: {A} Large-scale Dataset for Instance Segmentation in Aerial Images},
  booktitle    = {{IEEE} CVPRW},
  year         = {2019},
}

@inproceedings{DBLP:conf/nips/WangZMLZ21,
  author       = {Junjue Wang and others},
  title        = {LoveDA: {A} Remote Sensing Land-Cover Dataset for Domain Adaptive Semantic Segmentation},
  booktitle    = {NeurIPS},
  year         = {2021},
}

@article{DBLP:journals/tgrs/JiWL19,
  author       = {Shunping Ji and others},
  title        = {Fully Convolutional Networks for Multisource Building Extraction From an Open Aerial and Satellite Imagery Data Set},
  journal      = {{IEEE} TGRS},
  year         = {2019},
  doi          = {10.1109/TGRS.2018.2858817},
}

@inproceedings{DBLP:conf/igarss/MaggioriTCA17,
  author       = {Emmanuel Maggiori and others},
  title        = {Can semantic labeling methods generalize to any city? the inria aerial image labeling benchmark},
  booktitle    = {{IEEE} IGARSS},
  year         = {2017},
  doi          = {10.1109/IGARSS.2017.8127684},
}

@article{DBLP:journals/remotesensing/ChenS20,
  author       = {Hao Chen and others},
  title        = {A Spatial-Temporal Attention-Based Method and a New Dataset for Remote Sensing Image Change Detection},
  journal      = {Remote Sens.},
  year         = {2020},
  doi          = {10.3390/RS12101662},
}

@Article{isprs-archives-XLII-2-565-2018,
  author       = {M. A. Lebedev and others},
  title        = {CHANGE DETECTION IN REMOTE SENSING IMAGES USING CONDITIONAL ADVERSARIAL NETWORKS},
  journal      = {Int. Arch. Photogramm. Remote Sens. Spatial Inf. Sci.},
  year         = {2018},
  doi          = {10.5194/isprs-archives-XLII-2-565-2018},
}

@article{DBLP:journals/tgrs/ShiLLLWZ22,
  author       = {Qian Shi and others},
  title        = {A Deeply Supervised Attention Metric-Based Network and an Open Aerial Image Dataset for Remote Sensing Change Detection},
  journal      = {{IEEE} TGRS},
  year         = {2022},
  doi          = {10.1109/TGRS.2021.3085870},
}

@article{DBLP:journals/remotesensing/ShenLCWXYCLJ21,
  author       = {Li Shen and others},
  title        = {S2Looking: {A} Satellite Side-Looking Dataset for Building Change Detection},
  journal      = {Remote Sens.},
  year         = {2021},
  doi          = {10.3390/RS13245094},
}

@article{DBLP:journals/tgrs/YangXLDYPZ22,
  author       = {Kunping Yang and others},
  title        = {Asymmetric Siamese Networks for Semantic Change Detection in Aerial Images},
  journal      = {{IEEE} TGRS},
  year         = {2022},
  doi          = {10.1109/TGRS.2021.3113912},
}

@article{DBLP:journals/tgrs/YuJ22,
  author       = {Dawen Yu and others},
  title        = {A New Spatial-Oriented Object Detection Framework for Remote Sensing Images},
  journal      = {{IEEE} TGRS},
  year         = {2022},
  doi          = {10.1109/TGRS.2021.3127232},
}

@article{DBLP:journals/tgrs/HanYPP22,
  author       = {Yaqi Han and others},
  title        = {Fine-Grained Recognition for Oriented Ship Against Complex Scenes in Optical Remote Sensing Images},
  journal      = {{IEEE} TGRS},
  year         = {2022},
  doi          = {10.1109/TGRS.2021.3123666},
}

@article{DBLP:journals/tgrs/ChengWLXLYH22,
  author       = {Gong Cheng and others},
  title        = {Anchor-Free Oriented Proposal Generator for Object Detection},
  journal      = {{IEEE} TGRS},
  year         = {2022},
  doi          = {10.1109/TGRS.2022.3183022},
}

@article{DBLP:journals/digearth/YuanZZWLZ22,
  author       = {Panli Yuan and others},
  title        = {A transformer-based Siamese network and an open optical dataset for semantic change detection of remote sensing images},
  journal      = {Int. J. Digit. Earth},
  year         = {2022},
  doi          = {10.1080/17538947.2022.2111470},
}

@article{DBLP:journals/tgrs/ZhanXY23,
  author       = {Yang Zhan and others},
  title        = {{RSVG:} Exploring Data and Models for Visual Grounding on Remote Sensing Data},
  journal      = {{IEEE} TGRS},
  year         = {2023},
  doi          = {10.1109/TGRS.2023.3250471},
}

@article{DBLP:journals/tgrs/ZhangZGY24,
  author       = {Zilun Zhang and others},
  title        = {{RS5M} and GeoRSCLIP: {A} Large-Scale Vision-Language Dataset and a Large Vision-Language Model for Remote Sensing},
  journal      = {{IEEE} TGRS},
  year         = {2024},
  doi          = {10.1109/TGRS.2024.3449154},
}

@article{DBLP:journals/tgrs/YuanMXZ22,
  author       = {Zhenghang Yuan and others},
  title        = {Change Detection Meets Visual Question Answering},
  journal      = {{IEEE} TGRS},
  year         = {2022},
  doi          = {10.1109/TGRS.2022.3203314},
}

@inproceedings{DBLP:conf/igarss/SuiMZP23,
  author       = {Jialu Sui and others},
  title        = {{DTRN:} Dual Transformer Residual Network for Remote Sensing Super-Resolution},
  booktitle    = {{IEEE} IGARSS},
  year         = {2023},
  doi          = {10.1109/IGARSS52108.2023.10281785},
}

@article{DBLP:journals/tgrs/LeiSM22,
  author       = {Sen Lei and others},
  title        = {Transformer-Based Multistage Enhancement for Remote Sensing Image Super-Resolution},
  journal      = {{IEEE} TGRS},
  year         = {2022},
  doi          = {10.1109/TGRS.2021.3136190},
}

@article{DBLP:journals/tgrs/XiaoYJHJZ24,
  author       = {Yi Xiao and others},
  title        = {EDiffSR: An Efficient Diffusion Probabilistic Model for Remote Sensing Image Super-Resolution},
  journal      = {{IEEE} TGRS},
  year         = {2024},
  doi          = {10.1109/TGRS.2023.3341437},
}

@article{DBLP:journals/tgrs/ZhangLLGSJ24,
  author       = {Yan Zhang and others},
  title        = {{TCDM:} Effective Large-Factor Image Super-Resolution via Texture Consistency Diffusion},
  journal      = {{IEEE} TGRS},
  year         = {2024},
  doi          = {10.1109/TGRS.2024.3358913}
}

@article{DBLP:journals/tgrs/XiaoSYLSZ22,
  author       = {Yi Xiao. and others},
  title        = {Satellite Video Super-Resolution via Multiscale Deformable Convolution Alignment and Temporal Grouping Projection},
  journal      = {{IEEE} TGRS},
  year         = {2022},
  doi          = {10.1109/TGRS.2021.3107352}
}

@article{DBLP:journals/aeog/XiaoYHZSSWZ22,
  author       = {Yi Xiao and others},
  title        = {Space-time super-resolution for satellite video: {A} joint framework based on multi-scale spatial-temporal transformer},
  journal      = {Int. J. Appl. Earth Obs. Geoinf.},
  year         = {2022},
  doi          = {10.1016/J.JAG.2022.102731},
}

@article{DBLP:journals/tgrs/CuiDYC24,
  author       = {Mingpeng Cui and others},
  title        = {DOPNet: Dense Object Prediction Network for Multiclass Object Counting and Localization in Remote Sensing Images},
  journal      = {{IEEE} TGRS},
  year         = {2024},
  doi          = {10.1109/TGRS.2024.3349702},
}

@article{DBLP:journals/tgrs/WangZZWHWZ24,
  author       = {Huake Wang and others},
  title        = {Hierarchical Kernel Interaction Network for Remote Sensing Object Counting},
  journal      = {{IEEE} TGRS},
  year         = {2024},
  doi          = {10.1109/TGRS.2023.3348870},
}

@article{DBLP:journals/staeors/LiangZXJ24,
  author       = {Gongbo Liang and others},
  title        = {Unveiling Roadway Hazards: Enhancing Fatal Crash Risk Estimation Through Multiscale Satellite Imagery and Self-Supervised Cross-Matching},
  journal      = {{IEEE} JSTARS},
  year         = {2024},
  doi          = {10.1109/JSTARS.2023.3331438},
}

@article{TOLAN2024113888,
  author       = {Jamie Tolan and others},
  title        = {Very high resolution canopy height maps from RGB imagery using self-supervised vision transformer and convolutional decoder trained on aerial lidar},
  journal      = {Remote Sens. Environ.},
  year         = {2024},
  doi          = {10.1016/j.rse.2023.113888},
}

@article{DBLP:journals/tgrs/XiangL24,
  author       = {Shao Xiang and others},
  title        = {Remote Sensing Image Compression Based on High-Frequency and Low-Frequency Components},
  journal      = {{IEEE} TGRS},
  year         = {2024},
  doi          = {10.1109/TGRS.2023.3349306},
}

@inproceedings{DBLP:conf/cvpr/ZhuS022,
  author       = {Sijie Zhu and others},
  title        = {TransGeo: Transformer Is All You Need for Cross-view Image Geo-localization},
  booktitle    = {{IEEE/CVF} CVPR},
  year         = {2022},
  doi          = {10.1109/CVPR52688.2022.00123},
}

@article{DBLP:journals/tgrs/WangSQJLS0Q24,
  author       = {Zhen Wang and others},
  title        = {Sequence Matching for Image-Based UAV-to-Satellite Geolocalization},
  journal      = {{IEEE} TGRS},
  year         = {2024},
  doi          = {10.1109/TGRS.2024.3359605},
}

@article{DBLP:journals/tgrs/YangLZ24,
  author       = {Cong Yang and others},
  title        = {Bootstrapping Interactive Image-Text Alignment for Remote Sensing Image Captioning},
  journal      = {{IEEE} TGRS},
  year         = {2024},
  doi          = {10.1109/TGRS.2024.3359316},
}

@article{DBLP:journals/tgrs/MiaoHGJ24,
  author       = {Wang Miao and others},
  title        = {Hierarchical Feature Progressive Alignment Network for Remote Sensing Image Scene Classification in Multitarget Domain Adaptation},
  journal      = {{IEEE} TGRS},
  year         = {2024},
  doi          = {10.1109/TGRS.2023.3347618},
}

@article{DBLP:journals/tgrs/LiC0W24,
  author       = {Yangfan Li and others},
  title        = {Few-Shot Fine-Grained Classification With Rotation-Invariant Feature Map Complementary Reconstruction Network},
  journal      = {{IEEE} TGRS},
  year         = {2024},
  doi          = {10.1109/TGRS.2024.3361501},
}

@article{DBLP:journals/lgrs/DengXH22,
  author       = {Peifang Deng and others},
  title        = {When CNNs Meet Vision Transformer: {A} Joint Framework for Remote Sensing Scene Classification},
  journal      = {{IEEE} GRSL},
  year         = {2022},
  doi          = {10.1109/LGRS.2021.3109061},
}

@article{DBLP:journals/tgrs/WangLZTC23,
  author       = {Junjie Wang and others},
  title        = {Remote-Sensing Scene Classification via Multistage Self-Guided Separation Network},
  journal      = {{IEEE} TGRS},
  year         = {2023},
  doi          = {10.1109/TGRS.2023.3295797},
}

@article{AMOBOATENG2022569,
  author       = {Mark Amo-Boateng and others},
  title        = {Instance segmentation scheme for roofs in rural areas based on Mask R-CNN},
  journal      = {Egypt. J. Remote Sens. Space Sci.},
  year         = {2022},
  doi          = {10.1016/j.ejrs.2022.03.017},
}

@article{CHEN2023129,
  author       = {Shenglong Chen and others},
  title        = {Large-scale individual building extraction from open-source satellite imagery via super-resolution-based instance segmentation approach},
  journal      = {ISPRS J. Photogramm. Remote Sens.},
  year         = {2023},
}

@inproceedings{DBLP:conf/iccv/HeGDG17,
  author       = {Kaiming He and others},
  title        = {Mask {R-CNN}},
  booktitle    = {{IEEE} ICCV},
  year         = {2017},
  doi          = {10.1109/ICCV.2017.322},
}

@article{DBLP:journals/tgrs/LiCM24,
  author       = {Kaiyu Li and others},
  title        = {A New Learning Paradigm for Foundation Model-Based Remote-Sensing Change Detection},
  journal      = {{IEEE} TGRS},
  year         = {2024},
  doi          = {10.1109/TGRS.2024.3365825},
}

@article{WIELAND2023113452,
  author       = {Marc Wieland and others},
  title        = {Semantic segmentation of water bodies in very high-resolution satellite and aerial images},
  journal      = {Remote Sens. Environ.},
  year         = {2023},
  doi          = {https://doi.org/10.1016/j.rse.2023.113452},
}

@article{JIANG202410,
  author       = {Yiwen Jiang and others},
  title        = {Multi-branch reverse attention semantic segmentation network for building extraction},
  journal      = {Egypt. J. Remote Sens. Space Sci.},
  year         = {2024},
  doi          = {10.1016/j.ejrs.2023.12.003},
}

@article{xiao2023enhancing,
  author       = {Tao Xiao and others},
  title        = {Enhancing multiscale representations with transformer for remote sensing image semantic segmentation},
  journal      = {{IEEE} TGRS},
  year         = {2023},
}

@article{fu2024complementarity,
  author       = {Wei Fu and others},
  title        = {Complementarity-aware local-global feature fusion network for building extraction in remote sensing images},
  journal      = {{IEEE} TGRS},
  year         = {2024},
}

@article{dai2023radanet,
  author       = {Ling Dai and others},
  title        = {RADANet: Road augmented deformable attention network for road extraction from complex high-resolution remote-sensing images},
  journal      = {{IEEE} TGRS},
  year         = {2023},
}

@article{osco2023segment,
  author       = {Lucas Prado Osco and others},
  title        = {The segment anything model (sam) for remote sensing applications: From zero to one shot},
  journal      = {Int. J. Appl. Earth Obs. Geoinf.},
  year         = {2023},
}

@article{Wang31122022,
  author       = {Zhen Wang and others},
  title        = {Urban building extraction from high-resolution remote sensing imagery based on multi-scale recurrent conditional generative adversarial network},
  journal      = {GISci. Remote Sens.},
  year         = {2022},
  doi          = {10.1080/15481603.2022.2076382},
}

@article{zhang2021stagewise,
  author       = {Lefei Zhang and others},
  title        = {Stagewise unsupervised domain adaptation with adversarial self-training for road segmentation of remote-sensing images},
  journal      = {{IEEE} TGRS},
  year         = {2021},
}

@inproceedings{ayala2023diffusion,
  author       = {Christian Ayala and others},
  title        = {Diffusion models for remote sensing imagery semantic segmentation},
  booktitle    = {{IEEE} IGARSS},
  year         = {2023},
}

@ARTICLE{10416252,
  author       = {Yueqian Quan and others},
  title        = {ORSI Salient Object Detection via Progressive Semantic Flow and Uncertainty-Aware Refinement},
  journal      = {{IEEE} TGRS},
  year         = {2024},
  doi          = {10.1109/TGRS.2024.3359684},
}

@ARTICLE{9631225,
  author       = {Gongyang Li and others},
  title        = {Multi-Content Complementation Network for Salient Object Detection in Optical Remote Sensing Images},
  journal      = {{IEEE} TGRS},
  year         = {2022},
  doi          = {10.1109/TGRS.2021.3131221},
}

@ARTICLE{9474908,
  author       = {Xiaofei Zhou and others},
  title        = {Edge-Aware Multiscale Feature Integration Network for Salient Object Detection in Optical Remote Sensing Images},
  journal      = {{IEEE} TGRS},
  year         = {2022},
  doi          = {10.1109/TGRS.2021.3091312},
}

@ARTICLE{9791375,
  author       = {Qi Wang and others},
  title        = {Hybrid Feature Aligned Network for Salient Object Detection in Optical Remote Sensing Imagery},
  journal      = {{IEEE} TGRS},
  year         = {2022},
  doi          = {10.1109/TGRS.2022.3181062},
}

@article{Li18082022,
  author       = {Xungen Li and others},
  title        = {Ship detection of optical remote sensing image in multiple scenes},
  journal      = {Int. J. Remote Sens.},
  year         = {2022},
  doi          = {10.1080/01431161.2021.1931544},
}

@ARTICLE{10535900,
  author       = {Jie Feng and others},
  title        = {CFDRM: Coarse-to-Fine Dynamic Refinement Model for Weakly Supervised Moving Vehicle Detection in Satellite Videos},
  journal      = {{IEEE} TGRS},
  year         = {2024},
  doi          = {10.1109/TGRS.2024.3403868},
}

@ARTICLE{10990024,
  author       = {Jian Wang and others},
  title        = {Auto-Prompting SAM for Weakly Supervised Landslide Extraction},
  journal      = {{IEEE} GRSL},
  year         = {2025},
  doi          = {10.1109/LGRS.2025.3567600},
}

@article{DBLP:journals/staeors/YiLZL24,
  author       = {Hao Yi and others},
  title        = {Small Object Detection Algorithm Based on Improved YOLOv8 for Remote Sensing},
  journal      = {{IEEE} JSTARS},
  year         = {2024},
  doi          = {10.1109/JSTARS.2023.3339235},
}

@article{DBLP:journals/tgrs/ZhangWGZL23,
  author       = {Yunzuo Zhang and others},
  title        = {CFANet: Efficient Detection of {UAV} Image Based on Cross-Layer Feature Aggregation},
  journal      = {{IEEE} TGRS},
  year         = {2023},
  doi          = {10.1109/TGRS.2023.3273314},
}

@inproceedings{DBLP:conf/cvpr/DuHC023,
  author       = {Bowei Du and others},
  title        = {Adaptive Sparse Convolutional Networks with Global Context Enhancement for Faster Object Detection on Drone Images},
  booktitle    = {{IEEE/CVF} CVPR},
  year         = {2023},
  doi          = {10.1109/CVPR52729.2023.01291},
}

@article{DBLP:journals/tgrs/WangZCLZLDS22,
  author       = {Guanqun Wang and others},
  title        = {FSoD-Net: Full-Scale Object Detection From Optical Remote Sensing Imagery},
  journal      = {{IEEE} TGRS},
  year         = {2022},
  doi          = {10.1109/TGRS.2021.3064599},
}

@article{DBLP:journals/tgrs/ZhangLW23,
  author       = {Cong Zhang and others},
  title        = {CoF-Net: {A} Progressive Coarse-to-Fine Framework for Object Detection in Remote-Sensing Imagery},
  journal      = {{IEEE} TGRS},
  year         = {2023},
  doi          = {10.1109/TGRS.2022.3233881},
}

@article{DBLP:journals/tgrs/GaoLWCNL24,
  author       = {Tao Gao and others},
  title        = {Attention-Free Global Multiscale Fusion Network for Remote Sensing Object Detection},
  journal      = {{IEEE} TGRS},
  year         = {2024},
  doi          = {10.1109/TGRS.2023.3346041},
}

@article{DBLP:journals/tgrs/MingMZD22,
  author       = {Qi Ming and others},
  title        = {CFC-Net: {A} Critical Feature Capturing Network for Arbitrary-Oriented Object Detection in Remote-Sensing Images},
  journal      = {{IEEE} TGRS},
  year         = {2022},
  doi          = {10.1109/TGRS.2021.3095186},
}

@article{DBLP:journals/tgrs/ChengYLLXWYH22,
  author       = {Gong Cheng and others},
  title        = {Dual-Aligned Oriented Detector},
  journal      = {{IEEE} TGRS},
  year         = {2022},
  doi          = {10.1109/TGRS.2022.3149780},
}

@article{DBLP:journals/tgrs/HanDLX22,
  author       = {Jiaming Han and others},
  title        = {Align Deep Features for Oriented Object Detection},
  journal      = {{IEEE} TGRS},
  year         = {2022},
  doi          = {10.1109/TGRS.2021.3062048},
}

@article{DBLP:journals/tgrs/YaoCWLZXH23,
  author       = {Yanqing Yao and others},
  title        = {On Improving Bounding Box Representations for Oriented Object Detection},
  journal      = {{IEEE} TGRS},
  year         = {2023},
  doi          = {10.1109/TGRS.2022.3231340},
}

@article{DBLP:journals/tgrs/QianW0Y0H23,
  author       = {Xiaoliang Qian and others},
  title        = {Building a Bridge of Bounding Box Regression Between Oriented and Horizontal Object Detection in Remote Sensing Images},
  journal      = {{IEEE} TGRS},
  year         = {2023},
  doi          = {10.1109/TGRS.2023.3256373},
}

@article{DBLP:journals/tgrs/ZengCYLY24,
  author       = {Ying Zeng and others},
  title        = {{ARS-DETR:} Aspect Ratio-Sensitive Detection Transformer for Aerial Oriented Object Detection},
  journal      = {{IEEE} TGRS},
  year         = {2024},
  doi          = {10.1109/TGRS.2024.3364713},
}

@article{DBLP:journals/tgrs/WangZXZDTZ23,
  author       = {Di Wang and others},
  title        = {Advancing Plain Vision Transformer Toward Remote Sensing Foundation Model},
  journal      = {{IEEE} TGRS},
  year         = {2023},
  doi          = {10.1109/TGRS.2022.3222818},
}

@inproceedings{DBLP:conf/eccv/LiuZRLZYJLYSZZ24,
  author       = {Shilong Liu and others},
  title        = {Grounding {DINO:} Marrying {DINO} with Grounded Pre-training for Open-Set Object Detection},
  booktitle    = {ECCV},
  year         = {2024},
  doi          = {10.1007/978-3-031-72970-6\_3},
}

@inproceedings{DBLP:conf/iccv/KirillovMRMRGXW23,
  author       = {Alexander Kirillov and others},
  title        = {Segment Anything},
  booktitle    = {{IEEE/CVF} ICCV},
  year         = {2023},
  doi          = {10.1109/ICCV51070.2023.00371},
}

@article{DBLP:journals/tgrs/DongGL24,
  author       = {Zhe Dong and others},
  title        = {Generative ConvNet Foundation Model With Sparse Modeling and Low-Frequency Reconstruction for Remote Sensing Image Interpretation},
  journal      = {{IEEE} TGRS},
  year         = {2024},
  doi          = {10.1109/TGRS.2023.3348479},
}

@article{DBLP:journals/tgrs/SunWLZLHLRYCHYWLF23,
  author       = {Xian Sun and others},
  title        = {RingMo: {A} Remote Sensing Foundation Model With Masked Image Modeling},
  journal      = {{IEEE} TGRS},
  year         = {2023},
  doi          = {10.1109/TGRS.2022.3194732},
}

@article{DBLP:journals/tgrs/LiuCGZZYFZ24,
  author       = {Fan Liu and others},
  title        = {RemoteCLIP: {A} Vision Language Foundation Model for Remote Sensing},
  journal      = {{IEEE} TGRS},
  year         = {2024},
  doi          = {10.1109/TGRS.2024.3390838},
}

@inproceedings{DBLP:conf/icml/RadfordKHRGASAM21,
  author       = {Alec Radford and others},
  title        = {Learning Transferable Visual Models From Natural Language Supervision},
  booktitle    = {ICML},
  year         = {2021},
}

@article{Ge31122024,
  author       = {Jiayi Ge and others},
  title        = {Real-time identification of collapsed buildings triggered by natural disasters using a modified object-detection network with quasi-panchromatic images},
  journal      = {Eur. J. Remote Sens.},
  year         = {2024},
  doi          = {10.1080/22797254.2024.2318357},
}

@article{LONG2024318,
  author       = {Jiang Long and others},
  title        = {Semantic change detection using a hierarchical semantic graph interaction network from high-resolution remote sensing images},
  journal      = {ISPRS J. Photogramm. Remote Sens.},
  year         = {2024},
  doi          = {10.1016/j.isprsjprs.2024.04.012},
}

@article{ding2024joint,
  author       = {Lei Ding and others},
  title        = {Joint spatio-temporal modeling for semantic change detection in remote sensing images},
  journal      = {{IEEE} TGRS},
  year         = {2024},
}

@article{DBLP:journals/tgrs/DingZPTYB24,
  author       = {Lei Ding and others},
  title        = {Adapting Segment Anything Model for Change Detection in {VHR} Remote Sensing Images},
  journal      = {{IEEE} TGRS},
  year         = {2024},
  doi          = {10.1109/TGRS.2024.3368168},
}

@article{DBLP:journals/tgrs/LiZDD22,
  author       = {Qingyang Li and others},
  title        = {TransUNetCD: {A} Hybrid Transformer Network for Change Detection in Optical Remote-Sensing Images},
  journal      = {{IEEE} TGRS},
  year         = {2022},
  doi          = {10.1109/TGRS.2022.3169479},
}

@article{DBLP:journals/tgrs/TangFZFSLT24,
  author       = {Yingjie Tang and others},
  title        = {An Object Fine-Grained Change Detection Method Based on Frequency Decoupling Interaction for High-Resolution Remote Sensing Images},
  journal      = {{IEEE} TGRS},
  year         = {2024},
  doi          = {10.1109/TGRS.2023.3337816},
}

@article{DBLP:journals/tgrs/LiTLZDWZ23,
  author       = {Zhenglai Li and others},
  title        = {Lightweight Remote Sensing Change Detection With Progressive Feature Aggregation and Supervised Attention},
  journal      = {{IEEE} TGRS},
  year         = {2023},
  doi          = {10.1109/TGRS.2023.3241436},
}

@article{DBLP:journals/tgrs/ZhaoCZXBO24,
  author       = {Sijie Zhao and others},
  title        = {RS-Mamba for Large Remote Sensing Image Dense Prediction},
  journal      = {{IEEE} TGRS},
  year         = {2024},
  doi          = {10.1109/TGRS.2024.3425540},
}

@article{DBLP:journals/tgrs/ZhangCZCLZS24,
  author       = {Haotian Zhang and others},
  title        = {BiFA: Remote Sensing Image Change Detection With Bitemporal Feature Alignment},
  journal      = {{IEEE} TGRS},
  year         = {2024},
  doi          = {10.1109/TGRS.2024.3376673},
}

@article{DBLP:journals/tgrs/LiCDWWLP24,
  author       = {Zhi Li and others},
  title        = {STADE-CDNet: Spatial-Temporal Attention With Difference Enhancement-Based Network for Remote Sensing Image Change Detection},
  journal      = {{IEEE} TGRS},
  year         = {2024},
  doi          = {10.1109/TGRS.2024.3367948},
}

@article{DBLP:journals/tgrs/ChenQS22,
  author       = {Hao Chen and others},
  title        = {Remote Sensing Image Change Detection With Transformers},
  journal      = {{IEEE} TGRS},
  year         = {2022},
  doi          = {10.1109/TGRS.2021.3095166},
}

@article{DBLP:journals/tgrs/ChenSHXY24,
  author       = {Hongruixuan Chen. and others},
  title        = {ChangeMamba: Remote Sensing Change Detection With Spatiotemporal State Space Model},
  journal      = {{IEEE} TGRS},
  year         = {2024},
  doi          = {10.1109/TGRS.2024.3417253}
}

@article{lv2021land,
  author       = {Zhiyong Lv and others},
  title        = {Land cover change detection techniques: Very-high-resolution optical images: A review},
  journal      = {{IEEE} GRSM},
  year         = {2021},
}

@article{cheng2024methods,
  author       = {Jian Cheng and others},
  title        = {Methods and datasets on semantic segmentation for Unmanned Aerial Vehicle remote sensing images: A review},
  journal      = {ISPRS J. Photogramm. Remote Sens.},
  year         = {2024},
}

@article{li2022cloud,
  author       = {Zhiwei Li and others},
  title        = {Cloud and cloud shadow detection for optical satellite imagery: Features, algorithms, validation, and prospects},
  journal      = {ISPRS J. Photogramm. Remote Sens.},
  year         = {2022},
}

@article{zhang2024development,
  author       = {Chi Zhang and others},
  title        = {Development and application of ship detection and classification datasets: A review},
  journal      = {{IEEE} GRSM},
  year         = {2024},
}

@article{victor2024systematic,
  author       = {Brandon Victor and others},
  title        = {A systematic review of the use of Deep Learning in Satellite Imagery for Agriculture},
  journal      = {{IEEE} JSTARS},
  year         = {2024},
}

@article{li2024review,
  author       = {Jiangyun Li and others},
  title        = {A review of remote sensing image segmentation by deep learning methods},
  journal      = {Int. J. Digit. Earth},
  year         = {2024},
}

@article{li2022water,
  author       = {Yansheng Li and others},
  title        = {Water body classification from high-resolution optical remote sensing imagery: Achievements and perspectives},
  journal      = {ISPRS J. Photogramm. Remote Sens.},
  year         = {2022},
}

@ARTICLE{11119145,
  author       = {Dongshuo Yin and others},
  title        = {Remote sensing tuning: A survey},
  journal      = {Comput. Vis. Media},
  year         = {2025},
  doi          = {10.26599/CVM.2025.9450490},
}

@article{DBLP:journals/corr/abs-2503-22081,
  author       = {Ziyue Huang and others},
  title        = {A Survey on Remote Sensing Foundation Models: From Vision to Multimodality},
  journal      = {CoRR},
  year         = {2025},
  doi          = {10.48550/ARXIV.2503.22081},
}

@article{hosseiny2023beyond,
  author       = {Benyamin Hosseiny and others},
  title        = {Beyond supervised learning in remote sensing: A systematic review of deep learning approaches},
  journal      = {{IEEE} JSTARS},
  year         = {2023},
}

@article{wang2022empirical,
  author       = {Di Wang and others},
  title        = {An empirical study of remote sensing pretraining},
  journal      = {{IEEE} TGRS},
  year         = {2022},
}

@inproceedings{DBLP:conf/nips/KrizhevskySH12,
  author       = {Alex Krizhevsky and others},
  title        = {ImageNet Classification with Deep Convolutional Neural Networks},
  booktitle    = {NeurIPS},
  year         = {2012},
}

@inproceedings{DBLP:conf/iclr/DosovitskiyB0WZ21,
  author       = {Alexey Dosovitskiy and others},
  title        = {An Image is Worth 16x16 Words: Transformers for Image Recognition at Scale},
  booktitle    = {ICLR},
  year         = {2021},
}

@inproceedings{DBLP:conf/nips/VaswaniSPUJGKP17,
  author       = {Ashish Vaswani and others},
  title        = {Attention is All you Need},
  booktitle    = {NeurIPS},
  year         = {2017},
}

@Article{drones7060398,
  author       = {Zhengxin Zhang and others},
  title        = {A Review on Unmanned Aerial Vehicle Remote Sensing: Platforms, Sensors, Data Processing Methods, and Applications},
  journal      = {Drones},
  year         = {2023},
  doi          = {10.3390/drones7060398},
}

@article{DBLP:journals/tgrs/GaoLW21,
  author       = {Guangshuai Gao and others},
  title        = {Counting From Sky: {A} Large-Scale Data Set for Remote Sensing Object Counting and a Benchmark Method},
  journal      = {{IEEE} TGRS},
  year         = {2021},
  doi          = {10.1109/TGRS.2020.3020555},
}

@inproceedings{DBLP:conf/cvpr/WenDZHWBL21,
  author       = {Longyin Wen and others},
  title        = {Detection, Tracking, and Counting Meets Drones in Crowds: {A} Benchmark},
  booktitle    = {{IEEE} CVPR},
  year         = {2021},
  doi          = {10.1109/CVPR46437.2021.00772},
}

@article{DBLP:journals/network/SaadBC20,
  author       = {Walid Saad and others},
  title        = {A Vision of 6G Wireless Systems: Applications, Trends, Technologies, and Open Research Problems},
  journal      = {{IEEE} Netw.},
  year         = {2020},
  doi          = {10.1109/MNET.001.1900287},
}

@article{kim2024metasurface,
  author       = {Youngjin Kim and others},
  title        = {Metasurface folded lens system for ultrathin cameras},
  journal      = {Sci. Adv.},
  year         = {2024},
}

@article{rahman2019lossless,
  author       = {Md Atiqur Rahman and others},
  title        = {Lossless image compression techniques: A state-of-the-art survey},
  journal      = {Symmetry},
  year         = {2019},
}

@article{lee2023ipcc,
  author       = {Hoesung Lee and others},
  title        = {IPCC, 2023: Climate change 2023: Synthesis report, summary for policymakers},
  journal      = {IPCC, Geneva, Switzerland},
  year         = {2023},
}

@article{jutz2020copernicus,
  author       = {S Jutz and others},
  title        = {Copernicus: the european earth observation programme},
  journal      = {Rev. Teledetecci{\'o}n},
  year         = {2020},
}

@incollection{ramapriyan2009evolution,
  author       = {Hampapuram K. Ramapriyan and others},
  title        = {Evolution of the earth observing system (EOS) data and information system (EOSDIS)},
  booktitle    = {Standard-Based Data and Information Systems for Earth Observation},
  year         = {2009},
}

@book{kogan2019remote,
  author       = {Felix Kogan and others},
  title        = {Remote sensing for food security},
  year         = {2019},
}

@article{DBLP:journals/tgrs/HuangZGLW24,
  author       = {Ziyue Huang and others},
  title        = {Generic Knowledge Boosted Pretraining for Remote Sensing Images},
  journal      = {{IEEE} TGRS},
  year         = {2024},
  doi          = {10.1109/TGRS.2024.3354031},
}

@inproceedings{DBLP:conf/cvpr/RedmonDGF16,
  author       = {Joseph Redmon and others},
  title        = {You Only Look Once: Unified, Real-Time Object Detection},
  booktitle    = {2016 {IEEE} Conference on Computer Vision and Pattern Recognition,
                  {CVPR} 2016, Las Vegas, NV, USA, June 27-30, 2016},
  publisher    = {{IEEE} Computer Society},
  year         = {2016},
  doi          = {10.1109/CVPR.2016.91},
}

@article{konstantakos2025self,
  title={Self-supervised visual learning in the low-data regime: a comparative evaluation},
  author={Konstantakos, Sotirios and Cani, Jorgen and Mademlis, Ioannis and Chalkiadaki, Despina Ioanna and Asano, Yuki M and Gavves, Efstratios and Papadopoulos, Georgios Th},
  journal={Neurocomputing},
  volume={620},
  pages={129199},
  year={2025},
  publisher={Elsevier}
}

@article{alimisis2025advances,
  title={Advances in diffusion models for image data augmentation: A review of methods, models, evaluation metrics and future research directions},
  author={Alimisis, Panagiotis and Mademlis, Ioannis and Radoglou-Grammatikis, Panagiotis and Sarigiannidis, Panagiotis and Papadopoulos, Georgios Th},
  journal={Artificial Intelligence Review},
  volume={58},
  number={4},
  pages={112},
  year={2025},
  publisher={Springer}
}

@article{rodis2024multimodal,
  title={Multimodal explainable artificial intelligence: A comprehensive review of methodological advances and future research directions},
  author={Rodis, Nikolaos and Sardianos, Christos and Radoglou-Grammatikis, Panagiotis and Sarigiannidis, Panagiotis and Varlamis, Iraklis and Papadopoulos, Georgios Th},
  journal={IEEe Access},
  volume={12},
  pages={159794--159820},
  year={2024},
  publisher={IEEE}
}

@article{cani2026illicit,
  title={Illicit object detection in X-ray imaging using deep learning techniques: A comparative evaluation},
  author={Cani, Jorgen and Diou, Christos and Evangelatos, Spyridon and Argyriou, Vasileios and Radoglou-Grammatikis, Panagiotis and Sarigiannidis, Panagiotis and Varlamis, Iraklis and Papadopoulos, Georgios Th},
  journal={IEEE Access},
  year={2026},
  publisher={IEEE}
}
\end{document}